\documentclass[]{style}
\usepackage[toc,page,header]{appendix}
\usepackage{wrapfig}

\usepackage{booktabs}
\usepackage{multirow}
\usepackage[table]{xcolor}
\usepackage{colortbl}
\usepackage{array}
\usepackage{makecell}

\definecolor{headerblue}{RGB}{220,230,242}
\definecolor{groupgray}{RGB}{242,244,247}
\definecolor{semigreen}{RGB}{232,245,233}
\definecolor{bestred}{RGB}{255,235,238}
\definecolor{lightblue}{RGB}{245,249,255}

\usepackage{booktabs}
\usepackage{multirow}
\usepackage[table]{xcolor}
\usepackage{colortbl}
\usepackage{pifont}

\definecolor{headerblue}{RGB}{220,230,242}
\definecolor{groupgray}{RGB}{242,244,247}
\definecolor{semigreen}{RGB}{232,245,233}
\definecolor{bestred}{RGB}{255,235,238}
\definecolor{secondblue}{RGB}{232,240,254}
\definecolor{lightblue}{RGB}{245,249,255}
\usepackage{algorithm}
\usepackage{algpseudocode}

\usepackage{natbib}

\usepackage{CJKutf8}

\usepackage{xargs}

\usepackage{todonotes}

\usepackage{multirow}

\usepackage{cleveref}

\usepackage{amsmath}
\usepackage{amssymb}
\usepackage{mathtools}
\usepackage{amsthm}
\usepackage{dsfont}

\usepackage{svg}

\usepackage{mathrsfs}
\usepackage{adjustbox}
\usepackage{multicol}
\usepackage{tcolorbox}
\usepackage{changepage}
\usepackage{enumitem}
\usepackage{graphicx}
\usepackage{xcolor}
\usepackage{float}
\usepackage{threeparttable}
\usepackage{subcaption}
\usepackage{algorithm}
\usepackage{algpseudocode}
\usepackage{wrapfig}
\usepackage[table]{xcolor}
\usepackage{colortbl}
\usepackage{tabularx}
\usepackage{makecell}
\usepackage[dvipsnames]{xcolor}

\newcolumntype{g}{>{\columncolor{gray!10}}c}

\theoremstyle{plain}

\theoremstyle{definition}

\theoremstyle{remark}

\definecolor{catgray}{gray}{0.9}
\definecolor{skyblue}{rgb}{0.53,0.81,0.92}
\colorlet{skyblue!30}{skyblue!30!white}
\definecolor{customblue}{RGB}{70,130,180}

\renewcommand{\emph}[1]{\textit{#1}}

\usepackage{minitoc}
\usepackage{url}

\PassOptionsToPackage{table,xcdraw}{xcolor}
\usepackage{titletoc}
\usepackage{placeins}
\usepackage{pifont}

\definecolor{RowBlue}{HTML}{E9F2FB}
\definecolor{RowRed}{HTML}{F9EAEA}
\definecolor{Top1}{HTML}{50DB4B}
\definecolor{Top2}{HTML}{A5FFA2}
\definecolor{Top3}{HTML}{D9FFD9}
\definecolor{Sub1}{HTML}{EAB8B8}
\definecolor{Sub2}{HTML}{E4E4E4}

\title{Revisiting Chain-of-Thought Reasoning under Limited Supervision: Semi-supervised Chain-of-Thought Learning}

\author[1,3\dagger]{Hongyang He}
\author[2]{Jiuming Liu}
\author[1]{Victor Sanchez}

\affiliation[1]{University of Warwick}
\affiliation[2]{University of Cambridge}
\affiliation[3]{Manifolda.Ai}

\contribution[\dagger]{Corresponding author}

\abstract{
Chain-of-thought (CoT) reasoning has emerged as an effective approach for activating latent reasoning capabilities in large language models.
However, most existing CoT methods use reasoning chains mainly as inference-time prompts, while the generated reasoning traces are rarely reused as semi-supervised learning signals.
In this report, we define \textbf{Semi-supervised Chain-of-Thought Learning} and propose \textbf{Semi-CoT}, a simple framework that uses unlabeled questions to construct pseudo reasoning supervision.
Semi-CoT samples multiple pseudo-CoTs for each unlabeled question, estimates answer-level semantic entropy, and selects low-entropy reasoning chains as reliable pseudo-CoT demonstrations.
This extends the self-training view of CoT from inference-time refinement to semi-supervised pseudo-supervision.
Pilot experiments on AQuA, SVAMP, GSM8K, and MultiArith show that the entropy gate selects high-precision pseudo-CoTs, with pseudo-answer precision ranging from $91.36\%$ to $100\%$.
Semi-CoT also gives small gains on SVAMP and GSM8K, while AQuA shows negative transfer and MultiArith reaches a ceiling.
These results suggest that unlabeled questions can provide reliable pseudo reasoning signals, but their effective use still requires stronger demonstration selection or student training.
}

\begin{document}

\maketitle

\section{Introduction}
\label{sec:introduction}

Chain-of-thought (CoT) reasoning has become an effective way to elicit the reasoning ability of large language models (LLMs).
By asking a model to generate intermediate reasoning steps before producing the final answer, CoT improves performance on arithmetic, symbolic, and commonsense reasoning tasks~\cite{wei2022chain,kojima2022large}.
A series of follow-up studies further improve CoT through self-consistency, automatic demonstration construction, planning-based prompting, contrastive reasoning, and iterative re-reading~\cite{wang2022self,zhang2022automatic,wang2023plan,chia2023contrastive,xu2024re}.
Recent surveys also show that CoT has become a central component of reasoning with foundation models~\cite{chu2024navigate,sun2025survey}.
However, a common practice is still to use CoT only at inference time.
The model generates reasoning paths for a test question, uses them to reach a final answer, and then discards the generated reasoning traces.

This view leaves an important question underexplored:
can model-generated reasoning chains be used as semi-supervised learning signals?
In many reasoning tasks, high-quality CoT annotations are expensive.
Annotating a final answer already requires human effort, and writing a complete reasoning chain requires even more cost.
Moreover, different annotators may solve the same problem through different valid reasoning paths.
At the same time, unlabeled questions are often much easier to collect.
For example, math word problems, science questions, coding questions, and logic questions can exist in large quantities without human-written rationales~\cite{cobbe2021training,geva2021did}.
These unlabeled questions may still contain useful reasoning structures that can be discovered by an LLM.

Our motivation is closely related to self-training and semi-supervised learning.
Self-training uses model-generated pseudo-labels to exploit unlabeled data and has a long history in machine learning~\cite{scudder1965adaptive,lee2013pseudo,amini2025self}.
Modern semi-supervised learning further improves pseudo-labeling with entropy minimization, consistency regularization, confidence thresholding, curriculum pseudo-labeling, and teacher-student learning~\cite{grandvalet2004semi,tarvainen2017mean,sohn2020fixmatch,zhang2021flexmatch,yang2022survey}.
These methods show that unlabeled data can be useful when pseudo-labels are sufficiently reliable.
However, standard pseudo-labeling usually considers only the final label.
For reasoning tasks, the supervision signal should include not only the final answer but also the reasoning process that leads to it.

Recent work has observed that CoT reasoning and self-training share a similar structure~\cite{wu2024rethinking}.
In self-training, a model generates pseudo-labels for unlabeled samples and then uses them to improve learning.
In CoT reasoning, a model generates intermediate reasoning steps and uses them to improve prediction.
Both processes rely on model-generated information, and both aim to reduce uncertainty.
This connection suggests a natural extension:
if CoT can be viewed as a form of self-training at inference time, then model-generated CoTs should also be considered as pseudo-supervision at learning time.

Motivated by this observation, we define a new problem setting called \textbf{Semi-supervised Chain-of-Thought Learning}.
Given a small labeled question set and a larger unlabeled question set, the goal is to use unlabeled questions to generate reliable pseudo reasoning supervision.
Different from standard pseudo-labeling, the pseudo-supervision in this setting contains both a final answer and a reasoning chain:
\begin{equation}
x_u
\;\longrightarrow\;
(\hat r_u,\hat y_u),
\end{equation}
where $x_u$ is an unlabeled question, $\hat r_u$ is a pseudo reasoning chain, and $\hat y_u$ is a pseudo answer.
This formulation moves CoT from an inference-time prompting technique toward a semi-supervised learning signal.

However, directly using pseudo-CoTs is risky.
A teacher model may generate a wrong reasoning chain, a correct answer with an invalid derivation, or a plausible-looking reasoning path that leads to an incorrect answer.
Iterative reasoning may also introduce over-reasoning, where a previously correct answer becomes wrong after unnecessary additional reasoning~\cite{wu2024rethinking,nayab2024concise}.
Verifier-based methods show that checking reasoning quality is important for mathematical reasoning and CoT validation~\cite{cobbe2021training,ling2023deductive}.
Therefore, Semi-supervised CoT learning cannot simply generate CoTs on all unlabeled questions and use them as supervision.
It requires a reliability mechanism that can decide which pseudo-CoTs are trustworthy.

We use semantic entropy as a simple reliability signal.
Semantic entropy has been used to measure uncertainty over semantically distinct generations and to detect unreliable LLM outputs~\cite{farquhar2024detecting}.
For each unlabeled question, the model samples multiple reasoning paths and extracts their final answers.
If the sampled answers agree, the answer-level semantic entropy is low.
If the answers are inconsistent, the entropy is high.
We keep only low-entropy pseudo-CoTs and use them as reliable reasoning demonstrations.
This design follows the entropy-minimization view of semi-supervised learning~\cite{grandvalet2004semi} and the entropy-based view of CoT reasoning~\cite{wu2024rethinking}, but changes the role of semantic entropy:
instead of using it only as an inference-time stopping or uncertainty signal, we use it as a pseudo-CoT selection signal.

In this technical report, we implement a minimal version of Semi-supervised Chain-of-Thought Learning.
The method first splits a reasoning dataset into a small labeled set and a larger unlabeled set.
It then generates multiple pseudo-CoTs for each unlabeled question, computes semantic entropy over the extracted answers, and stores accepted pseudo-CoTs in a pseudo reasoning bank.
During inference, the model uses selected pseudo-CoTs as demonstrations for solving test questions.
This implementation is deliberately simple.
It does not fine-tune model parameters, but it directly tests whether unlabeled questions can provide useful pseudo reasoning supervision.

We conduct pilot experiments on AQuA, SVAMP, GSM8K, and MultiArith, which are representative arithmetic reasoning benchmarks.
The results show that the entropy gate selects high-precision pseudo-CoTs across datasets.
The pseudo-answer precision reaches $97.14\%$ on AQuA, $96.30\%$ on SVAMP, $91.36\%$ on GSM8K, and $100\%$ on MultiArith.
In terms of accuracy, Semi-CoT improves over zero-shot CoT by $2$ points on SVAMP and $1$ point on GSM8K in the current pilot setting.
AQuA shows mild negative transfer, and MultiArith reaches a ceiling where all compared methods obtain $100\%$ accuracy.
These results suggest that semantic entropy is useful for selecting reliable pseudo-CoTs, but also show that reliable pseudo-CoTs must be used carefully.
Random demonstrations can be unstable, and stronger retrieval or student training may be needed for larger gains.

Our contributions are summarized as follows:
\begin{itemize}
    \item We define \textbf{Semi-supervised Chain-of-Thought Learning}, a new setting that uses unlabeled questions to construct pseudo reasoning supervision.
    \item We propose a minimal \textbf{Semi-CoT} pipeline that generates multiple pseudo-CoTs, filters them with semantic entropy, and uses accepted pseudo-CoTs as reasoning demonstrations.
    \item We provide pilot evidence that entropy-based pseudo-CoT selection yields high pseudo-answer precision across multiple reasoning datasets.
    \item We identify an important limitation of the current prompt-level implementation: pseudo-CoT quality alone is not sufficient, and demonstration relevance is necessary to avoid negative transfer.
\end{itemize}
\section{Related Work and Preliminaries}
\label{sec:preliminaries}

\subsection{Chain-of-Thought Reasoning}

Chain-of-thought (CoT) reasoning improves the reasoning ability of large language models by asking them to generate intermediate steps before producing the final answer~\cite{wei2022chain}.
Early CoT prompting mainly relies on manually written demonstrations, where each example contains a question, a reasoning chain, and a final answer.
Zero-shot CoT removes the need for manually written demonstrations by using a general instruction such as ``Let's think step by step''~\cite{kojima2022large}.
Self-consistency further improves CoT by sampling multiple reasoning paths and selecting the answer with majority voting~\cite{wang2022self}, which is related to the broader idea that voting and margin effects can improve prediction reliability~\cite{bartlett1998boosting}.
Follow-up methods improve CoT from different directions, including automatic demonstration construction~\cite{zhang2022automatic}, planning-based prompting~\cite{wang2023plan}, contrastive CoT prompting~\cite{chia2023contrastive}, iterative re-reading~\cite{xu2024re}, and output-length control for reasoning efficiency~\cite{nayab2024concise}.
Recent surveys further summarize the progress of CoT and foundation-model reasoning~\cite{chu2024navigate,sun2025survey}.

These studies show that reasoning traces can help LLMs solve arithmetic, symbolic, and commonsense reasoning tasks.
However, most existing CoT methods use reasoning only during inference.
The model generates reasoning paths for a test question, uses them to reach a final answer, and then discards them.
As a result, the generated reasoning traces are not reused as learning signals.
This limits the role of CoT to a prompt-level inference strategy.
In contrast, Semi-CoT studies whether generated reasoning chains can be converted into pseudo-supervision for unlabeled questions.

\subsection{Reasoning Benchmarks and Verification}

CoT reasoning is commonly evaluated on arithmetic, commonsense, symbolic, and multi-step question answering tasks.
Arithmetic word problem datasets provide an important testbed because they require both numerical computation and intermediate reasoning.
Representative datasets include arithmetic word problem benchmarks based on verb categorization~\cite{hosseini2014learning}, algebraic equation parsing~\cite{koncel2015parsing}, rationale generation for algebraic reasoning~\cite{ling2017program}, general arithmetic word problems~\cite{roy2015solving}, and robustness-oriented math word problem evaluation~\cite{patel2021nlp}.
GSM8K further provides grade-school math problems with natural language solutions and has become a standard benchmark for verifier-based mathematical reasoning~\cite{cobbe2021training}.
Beyond arithmetic reasoning, benchmarks such as StrategyQA and CommonsenseQA evaluate implicit reasoning and commonsense question answering~\cite{geva2021did,talmor2019commonsenseqa}.
These benchmarks motivate our evaluation on AQuA, SVAMP, GSM8K, and MultiArith, where the goal is not only to predict the final answer but also to test whether unlabeled questions can provide useful pseudo reasoning supervision.

Reasoning quality is not fully captured by final-answer accuracy.
A model can produce the correct answer for the wrong reason, or generate a plausible-looking chain that leads to an incorrect answer.
Verifier-based methods address this issue by training models to check mathematical solutions or verify deductive reasoning steps~\cite{cobbe2021training,ling2023deductive}.
Recent evaluations of stronger reasoning models also emphasize that reasoning reliability, hallucination control, and verification remain open challenges~\cite{zhong2024evaluation}.
These observations motivate our reliability-oriented treatment of pseudo-CoTs.
Instead of assuming that every generated reasoning chain is useful, Semi-CoT first filters pseudo-CoTs before using them as demonstrations or future training targets.

\subsection{Self-Training and Semi-Supervised Learning}

Semi-supervised learning uses a small labeled set and a large unlabeled set to improve model performance.
A common strategy is self-training, where a teacher model or the current model generates pseudo-labels for unlabeled samples, and the student is trained with both labeled and pseudo-labeled data~\cite{scudder1965adaptive,lee2013pseudo,amini2025self,yang2022survey}.
Let $\mathcal{D}_l=\{(x_i,y_i)\}_{i=1}^{N_l}$ denote the labeled set and $\mathcal{D}_u=\{x_j\}_{j=1}^{N_u}$ denote the unlabeled set.
A standard pseudo-labeling method first predicts $\hat y_j$ for each unlabeled sample $x_j$, and then trains the model with both $(x_i,y_i)$ and $(x_j,\hat y_j)$.

The success of self-training depends strongly on pseudo-label reliability.
Incorrect pseudo-labels can mislead the model and cause error accumulation.
Therefore, many semi-supervised methods use confidence thresholding, entropy minimization, consistency regularization, curriculum pseudo-labeling, uncertainty-aware selection, or teacher-student averaging to reduce pseudo-label noise~\cite{grandvalet2004semi,tarvainen2017mean,sohn2020fixmatch,zhang2021flexmatch,mukherjee2020uncertainty}.
Other works study debiasing and confidence regularization in self-training, especially when pseudo-label distributions are biased or over-confident~\cite{chen2022debiased,zou2019confidence}.
Virtual adversarial training and consistency-based regularization also show that unlabeled data can improve robustness when predictions are encouraged to be stable under perturbations~\cite{miyato2018virtual}.
Theoretical studies further explain how self-training can turn weak learners into strong learners under mixture-model assumptions~\cite{frei2022self}.

These ideas are useful for classification, where the pseudo-supervision is usually a discrete class label.
Recent semi-supervised vision studies also show that pseudo-label reliability remains central in fine-grained recognition, long-tailed recognition, and robust visual learning~\cite{A38,A39,A40,A41,he2025semi,A15,he2026newton}.
Related vision works on deepfake detection and road damage detection further reflect the broader need for robust supervision under noisy or limited labels~\cite{A37,wan2022yolo}.
However, reasoning tasks require a richer form of pseudo-supervision.
The target is not only the final answer, but also the reasoning process that leads to the answer.
Semi-CoT therefore extends pseudo-labeling from answer-only supervision to reasoning-aware pseudo-supervision:
\begin{equation}
x_u
\;\longrightarrow\;
(\hat r_u,\hat y_u),
\label{eq:pseudo_cot}
\end{equation}
where $x_u$ is an unlabeled question, $\hat r_u$ is a pseudo reasoning chain, and $\hat y_u$ is a pseudo answer.

\subsection{CoT from the Perspective of Self-Training}

Recent work has pointed out that CoT reasoning and self-training share a similar structure~\cite{wu2024rethinking}.
Both methods rely on model-generated information.
In self-training, the model generates pseudo-labels.
In CoT reasoning, the model generates reasoning traces.
Both methods also use an iterative process to reduce uncertainty and improve predictions.
This connection is important because reduce uncertainty and improve predictions.
This connection is it suggests that CoT should not only be interpreted as an inference-time reasoning trick.
It can also be interpreted as a form of model-generated supervision.

This connection gives a direct motivation for Semi-CoT.
If CoT can be interpreted as a self-training-like process at inference time, then generated CoTs should not only be used for test-time refinement.
They can also be treated as pseudo-supervision for unlabeled questions.
In this way, Semi-CoT moves from
\begin{equation}
\text{CoT as inference-time reasoning}
\end{equation}
to
\begin{equation}
\text{CoT as semi-supervised reasoning supervision}.
\end{equation}

This shift creates a new problem.
A generated CoT may be correct, partially correct, or completely wrong.
A model may also produce a correct answer with an invalid reasoning chain.
Moreover, iterative reasoning can introduce over-reasoning, where extra reasoning steps change a correct answer into an incorrect one~\cite{wu2024rethinking,nayab2024concise}.
Therefore, Semi-CoT needs a reliability mechanism before using pseudo-CoTs as supervision.

\subsection{Semantic Entropy and Uncertainty Estimation}

Semantic entropy measures uncertainty over semantically distinct answers.
Unlike token-level entropy, semantic entropy groups different surface forms that express the same meaning and then computes uncertainty over semantic clusters.
This makes it suitable for LLM generations, where the same answer may be expressed in many different forms~\cite{farquhar2024detecting}.
In CoT reasoning, a model can sample multiple reasoning paths for the same question.
Each path produces a final answer.
If these answers agree, the model has low answer-level uncertainty.
If they disagree, the model has high uncertainty.

For an unlabeled question $x_u$, we sample $K$ reasoning paths and obtain answer candidates $\{\hat y_u^k\}_{k=1}^{K}$.
After answer normalization, the candidates are grouped by answer semantics.
Let $p_c$ denote the empirical probability of answer group $c$.
We define the normalized semantic entropy as
\begin{equation}
H_{\mathrm{sem}}(x_u)
=
-\frac{1}{\log K}
\sum_{c} p_c \log p_c .
\label{eq:semantic_entropy}
\end{equation}
When all sampled answers agree, $H_{\mathrm{sem}}(x_u)=0$.
When the sampled answers are diverse, $H_{\mathrm{sem}}(x_u)$ becomes larger.

Semi-CoT uses this entropy as a pseudo-CoT selection signal.
A pseudo-CoT is accepted only when
\begin{equation}
H_{\mathrm{sem}}(x_u) \leq \delta,
\label{eq:entropy_gate}
\end{equation}
where $\delta$ is a predefined threshold.
The accepted pseudo answer is the majority answer, and the accepted pseudo reasoning chain is chosen from the sampled reasoning paths that lead to this majority answer.
This design treats low-entropy agreement as a simple reliability signal.
It is also connected to entropy minimization in semi-supervised learning~\cite{grandvalet2004semi}, uncertainty-aware self-training~\cite{mukherjee2020uncertainty}, and semantic-entropy-based uncertainty estimation for LLM outputs~\cite{farquhar2024detecting}.

\subsection{Retrieval and Demonstration Relevance}

Using pseudo-CoTs as demonstrations introduces another issue: reliability and relevance are different.
A pseudo-CoT may be correct for its own question but irrelevant to a new test question.
Randomly selected demonstrations may therefore introduce negative transfer.
This motivates retrieval-based demonstration selection.
Retrieval-augmented reasoning has been studied as a way to provide external or example-based context to LLMs~\cite{liu2024much}.
Classical lexical similarity measures, such as Jaccard coefficient and TF-IDF-style matching, provide simple ways to retrieve examples based on surface similarity.
However, lexical similarity may not fully capture reasoning similarity.
Two questions may share few words but require the same reasoning pattern, while two lexically similar questions may require different solution strategies.

In this report, we use TF-IDF retrieval only as a lightweight pilot baseline.
The goal is not to solve demonstration relevance completely, but to separate two factors:
pseudo-CoT reliability, controlled by the entropy gate, and pseudo-CoT relevance, controlled by the demonstration selection strategy.
Future Semi-CoT methods can use stronger semantic retrieval, answer-type matching, reasoning-template matching, verifier-based ranking, or parametric student training.

\subsection{Problem Setup of Semi-CoT}

We define \textbf{Semi-supervised Chain-of-Thought Learning} as follows.
Given a small labeled question set $\mathcal{D}_l$ and a larger unlabeled question set $\mathcal{D}_u$, the goal is to use unlabeled questions to construct reliable pseudo reasoning supervision.
The labeled set contains questions and gold answers.
The unlabeled set contains only questions during pseudo-CoT generation and selection.
Gold answers from the unlabeled set are not used for filtering or prompt construction.
They are used only for offline evaluation.

A general Semi-CoT method contains three steps.
First, it generates multiple candidate pseudo-CoTs for each unlabeled question.
Second, it estimates the reliability of these pseudo-CoTs.
Third, it uses reliable pseudo-CoTs as demonstrations or training targets.
The minimal implementation in this report follows the demonstration-based version.
It stores accepted pseudo-CoTs in a pseudo reasoning bank and retrieves examples from this bank during inference.

Formally, the pseudo reasoning bank is
\begin{equation}
\mathcal{B}
=
\{(x_u,\hat r_u,\hat y_u,w_u)
\mid
x_u\in\mathcal{D}_u,\;
H_{\mathrm{sem}}(x_u)\leq\delta
\},
\label{eq:pseudo_bank}
\end{equation}
where $w_u=1-H_{\mathrm{sem}}(x_u)$ is the pseudo-CoT reliability weight.
In the current prompt-level implementation, $w_u$ is stored for analysis and future training, but the inference prompt uses accepted pseudo-CoTs as demonstrations without weighted optimization.

\subsection{Scope of This Report}

This report focuses on the minimal setting needed to validate the idea of Semi-CoT.
The current method does not fine-tune LLM parameters.
Instead, it tests whether unlabeled questions can provide reliable pseudo-CoT demonstrations.
This is a first step toward a fuller training-based Semi-CoT framework, where accepted pseudo-CoTs can be used to train a student model by minimizing a supervised loss on $\mathcal{D}_l$ and a pseudo-CoT loss on $\mathcal{D}_u$.
A future parametric objective can be written as
\begin{equation}
\mathcal{L}
=
\mathcal{L}_{\mathrm{sup}}
+
\lambda_u
\mathbb{E}_{x_u\in\mathcal{D}_u}
\left[
w_u
\mathcal{L}_{\mathrm{pseudo\text{-}CoT}}
(x_u,\hat r_u,\hat y_u)
\right].
\label{eq:future_objective}
\end{equation}
This training-based extension is not the focus of the present implementation.
The goal of this report is to establish the problem setting and provide initial evidence that semantic entropy can select reliable pseudo-CoT signals from unlabeled questions.
\section{Method}
\label{sec:method}

\subsection{Overview}

We propose \textbf{Semi-CoT}, a minimal framework for semi-supervised chain-of-thought learning.
The goal is to use unlabeled questions as a source of pseudo reasoning supervision.
Given a small labeled set and a larger unlabeled set, Semi-CoT first generates multiple reasoning paths for each unlabeled question.
It then estimates the reliability of the generated pseudo-CoTs by measuring answer-level semantic entropy.
Only low-entropy pseudo-CoTs are accepted into a pseudo reasoning bank.
During inference, the model uses selected pseudo-CoTs from this bank as demonstrations for solving new questions.

This implementation is intentionally simple.
It does not update the parameters of the LLM.
Instead, it tests the first necessary condition of Semi-supervised CoT learning:
whether unlabeled questions can provide reliable pseudo reasoning traces.
A full training-based version can later use the same pseudo-CoT bank to fine-tune a student model.

\subsection{Semi-Supervised Reasoning Setup}

Let $\mathcal{D}=\{(x_i,y_i)\}_{i=1}^{N}$ be a reasoning dataset, where $x_i$ is a question and $y_i$ is the gold final answer.
We split $\mathcal{D}$ into a labeled set and an unlabeled set:
\begin{equation}
\mathcal{D}_l=\{(x_i,y_i)\}_{i=1}^{N_l},
\qquad
\mathcal{D}_u=\{x_j\}_{j=1}^{N_u}.
\end{equation}
Only the labeled set $\mathcal{D}_l$ can use gold answers during pseudo-CoT construction.
For the unlabeled set $\mathcal{D}_u$, gold answers are hidden during pseudo-CoT generation and filtering.
They are used only for offline diagnosis, such as pseudo-answer precision.

For each unlabeled question $x_u\in\mathcal{D}_u$, Semi-CoT aims to construct a pseudo reasoning pair:
\begin{equation}
x_u
\;\longrightarrow\;
(\hat r_u,\hat y_u),
\end{equation}
where $\hat r_u$ is a generated reasoning chain and $\hat y_u$ is the generated final answer.
The main problem is to decide whether $(\hat r_u,\hat y_u)$ is reliable enough to be used as pseudo-supervision.

\subsection{Pseudo-CoT Generation}

For each unlabeled question $x_u$, we sample $K$ chain-of-thought outputs from the model.
Each output contains a reasoning trace and a final answer:
\begin{equation}
(r_u^k,\hat y_u^k)
=
\mathrm{LLM}(x_u,p_{\mathrm{cot}};\epsilon_k),
\qquad
k=1,\ldots,K,
\end{equation}
where $p_{\mathrm{cot}}$ is the CoT trigger and $\epsilon_k$ denotes sampling randomness.
In practice, $p_{\mathrm{cot}}$ can be a standard prompt such as ``Let's think step by step.''

After generation, we parse each output and extract the final answer.
This gives an answer set:
\begin{equation}
\hat A_u
=
\{\hat y_u^1,\hat y_u^2,\ldots,\hat y_u^K\}.
\end{equation}
The extracted answers are normalized before comparison.
For numerical datasets, formatting symbols and redundant spaces are removed.
For multiple-choice datasets, answer options such as A, B, C, D, and E are preserved.

\subsection{Semantic-Entropy Reliability Gate}

A pseudo-CoT should be more reliable when independently sampled reasoning paths lead to the same answer.
Therefore, Semi-CoT uses answer-level semantic entropy as the reliability signal.
Let $\mathcal{C}_u$ be the set of normalized answer groups from $\hat A_u$.
For each answer group $c\in\mathcal{C}_u$, let $p_c$ be its empirical frequency:
\begin{equation}
p_c
=
\frac{1}{K}
\sum_{k=1}^{K}
\mathbb{I}[\hat y_u^k=c].
\end{equation}
The normalized semantic entropy is defined as
\begin{equation}
H_{\mathrm{sem}}(x_u)
=
-\frac{1}{\log K}
\sum_{c\in\mathcal{C}_u}
p_c\log p_c.
\label{eq:semicot_entropy}
\end{equation}
When all sampled answers are the same, $H_{\mathrm{sem}}(x_u)=0$.
When sampled answers disagree, $H_{\mathrm{sem}}(x_u)$ becomes larger.

Semi-CoT accepts a pseudo-CoT only if its entropy is below a threshold:
\begin{equation}
m_u
=
\mathbb{I}
\left[
H_{\mathrm{sem}}(x_u)\leq \delta
\right],
\label{eq:semicot_gate}
\end{equation}
where $\delta$ controls the strictness of pseudo-CoT selection.
A smaller $\delta$ keeps fewer but more consistent pseudo-CoTs.
A larger $\delta$ accepts more pseudo-CoTs but may introduce more noise.

For each accepted unlabeled question, the pseudo answer is chosen by majority voting:
\begin{equation}
\hat y_u
=
\arg\max_{c\in\mathcal{C}_u} p_c.
\end{equation}
The pseudo reasoning chain $\hat r_u$ is chosen from the generated paths whose extracted answer equals $\hat y_u$.
If multiple reasoning paths lead to the same majority answer, we choose the shortest valid path.
This reduces prompt length and avoids unnecessary reasoning verbosity.

\subsection{Pseudo Reasoning Bank}

All accepted pseudo-CoTs are stored in a pseudo reasoning bank:
\begin{equation}
\mathcal{B}
=
\left\{
(x_u,\hat r_u,\hat y_u,w_u)
\mid
x_u\in\mathcal{D}_u,\;
H_{\mathrm{sem}}(x_u)\leq\delta
\right\},
\label{eq:semicot_bank}
\end{equation}
where the reliability weight is
\begin{equation}
w_u = 1-H_{\mathrm{sem}}(x_u).
\end{equation}
In the current prompt-level implementation, $w_u$ is stored for analysis and future training.
It is not used as a loss weight because we do not fine-tune model parameters in this report.

For labeled questions, Semi-CoT can also construct a small set of verified CoT demonstrations.
For each labeled example $(x_i,y_i)\in\mathcal{D}_l$, the model samples several CoT paths.
A generated CoT is kept only if its extracted answer matches the gold answer $y_i$.
If no generated path matches the gold answer, we keep an answer-only fallback target.
This makes the labeled part of the bank consistent with the available gold supervision.

\subsection{Demonstration Selection}

At inference time, Semi-CoT selects $M$ examples from the pseudo reasoning bank and places them before the test question.
We study two simple strategies.

\textbf{Random selection.}
The first strategy randomly samples $M$ pseudo-CoTs from $\mathcal{B}$ with a fixed random seed.
This gives a basic test of whether pseudo-CoTs are useful as demonstrations.
However, random selection can introduce irrelevant demonstrations and may cause negative transfer.

\textbf{TF-IDF retrieval.}
The second strategy selects pseudo-CoTs that are lexically similar to the test question.
We implement a lightweight TF-IDF retriever without extra dependencies.
For a test question $x$, each candidate pseudo question $x_u$ receives a similarity score:
\begin{equation}
s(x,x_u)
=
\cos
\left(
\phi_{\mathrm{tfidf}}(x),
\phi_{\mathrm{tfidf}}(x_u)
\right),
\end{equation}
where $\phi_{\mathrm{tfidf}}(\cdot)$ is the TF-IDF vector.
The top-$M$ pseudo-CoTs are used as demonstrations.
If the test question itself appears in the pseudo bank, it is removed before retrieval.
This prevents data leakage.

\subsection{Semi-CoT Inference}

Given a test question $x$, Semi-CoT builds a demonstration used as demonstrations.
If prompt from selected pseudo-CoTs:
\begin{equation}
\mathcal{P}(x)
=
\left[
(x_{d_1},\hat r_{d_1},\hat y_{d_1}),
\ldots,
(x_{d_M},\hat r_{d_M},\hat y_{d_M}),
x
\right].
\end{equation}
The model then generates a reasoning chain and final answer:
\begin{equation}
(r,\hat y)
=
\mathrm{LLM}(\mathcal{P}(x)).
\end{equation}
The final answer $\hat y$ is extracted with the same answer parser used during pseudo-CoT generation.
Accuracy is computed by comparing $\hat y$ with the gold answer of the test question.

This inference process differs from standard zero-shot CoT.
Zero-shot CoT only uses a general reasoning trigger.
Semi-CoT augments the prompt with pseudo-CoT demonstrations mined from unlabeled questions.
Thus, the unlabeled set affects the final prediction through the pseudo reasoning bank.

The overall procedure is summarized in Algorithm~\ref{alg:semicot}.

\begin{algorithm}[t]
\caption{Semi-CoT: Semi-supervised Chain-of-Thought Learning}
\label{alg:semicot}
\begin{algorithmic}[1]
\Require Labeled set $\mathcal{D}_l$, unlabeled set $\mathcal{D}_u$, sampling number $K$, entropy threshold $\delta$, number of demonstrations $M$
\State Initialize pseudo reasoning bank $\mathcal{B}\leftarrow\emptyset$
\For{each unlabeled question $x_u\in\mathcal{D}_u$}
    \State Sample $K$ CoT outputs $\{(r_u^k,\hat y_u^k)\}_{k=1}^{K}$
    \State Normalize extracted answers $\{\hat y_u^k\}_{k=1}^{K}$
    \State Compute semantic entropy $H_{\mathrm{sem}}(x_u)$ by Eq.~\eqref{eq:semicot_entropy}
    \If{$H_{\mathrm{sem}}(x_u)\leq\delta$}
        \State Select majority answer $\hat y_u$
        \State Select a valid reasoning chain $\hat r_u$ that leads to $\hat y_u$
        \State Set reliability weight $w_u=1-H_{\mathrm{sem}}(x_u)$
        \State Add $(x_u,\hat r_u,\hat y_u,w_u)$ to $\mathcal{B}$
    \EndIf
\EndFor
\For{each test question $x$}
    \State Select $M$ demonstrations from $\mathcal{B}$ by random sampling or TF-IDF retrieval
    \State Build the Semi-CoT prompt with selected demonstrations and $x$
    \State Generate reasoning and final answer with the LLM
    \State Extract the final answer and evaluate correctness
\EndFor
\end{algorithmic}
\end{algorithm}

\subsection{Discussion}

The current Semi-CoT implementation is a prompt-level semi-supervised method.
It validates whether unlabeled questions can produce reliable pseudo reasoning signals.
It also exposes a key limitation:
high pseudo-CoT precision does not always guarantee better accuracy.
If demonstrations are irrelevant to the test question, they can still cause negative transfer.
Therefore, pseudo-CoT selection and demonstration selection are two different problems.
The entropy gate controls pseudo-CoT reliability, while retrieval controls pseudo-CoT relevance.

A fuller version of Semi-CoT can train a student model with accepted pseudo-CoTs.
In that case, the pseudo reasoning bank can be used to optimize
\begin{equation}
\mathcal{L}
=
\mathcal{L}_{\mathrm{sup}}
+
\lambda_u
\mathbb{E}_{x_u\in\mathcal{D}_u}
\left[
w_u
\mathcal{L}_{\mathrm{pseudo\text{-}CoT}}
(x_u,\hat r_u,\hat y_u)
\right].
\end{equation}
This training-based extension is left for future experiments.
The present report focuses on the minimal setting needed to define the task and test the reliability of entropy-filtered pseudo-CoTs.

\section{Theoretical Analysis}
\label{sec:theory}

This section provides a theoretical analysis of Semi-CoT from the perspective of pseudo-CoT uncertainty.
Our goal is not to prove that every low-entropy pseudo-CoT is correct.
Instead, we analyze why semantic entropy can be used as a reasonable reliability signal for pseudo-CoT selection.
The analysis follows the self-training view of CoT reasoning.
In standard self-training, unlabeled samples are useful only when their pseudo-labels are reliable.
In Semi-CoT, unlabeled questions are useful only when their generated pseudo reasoning chains and pseudo answers are reliable.
Therefore, the key question is whether the uncertainty of generated answers can indicate the reliability of pseudo-CoT supervision.

\subsection{Uncertainty in Pseudo-CoT Generation}

Let $x$ denote an unlabeled question.
Given a CoT prompt $p$ and a sampling temperature $\tau$, an LLM defines a distribution over reasoning--answer pairs:
\begin{equation}
(r,a) \sim P_{\mathrm{LLM}}(r,a\mid x,p,\tau),
\end{equation}
where $r$ is a reasoning chain and $a$ is the final answer extracted from the generated output.
For each unlabeled question, Semi-CoT samples $K$ reasoning--answer pairs:
\begin{equation}
\{(r^k,a^k)\}_{k=1}^{K}.
\end{equation}
The answers are then normalized and grouped into semantic answer clusters.
Let $\mathcal{C}(x)=\{C_1,\ldots,C_m\}$ be the set of answer clusters for question $x$.
Each cluster $C_i$ contains answers with the same semantics.
Let $p_i$ be the empirical probability of cluster $C_i$:
\begin{equation}
p_i
=
\frac{1}{K}
\sum_{k=1}^{K}
\mathbb{I}[a^k\in C_i],
\qquad
\sum_{i=1}^{m}p_i=1.
\end{equation}
The normalized semantic entropy of $x$ is
\begin{equation}
H_{\mathrm{sem}}(x)
=
-\frac{1}{\log K}
\sum_{i=1}^{m}p_i\log p_i .
\label{eq:theory_semantic_entropy}
\end{equation}
When all sampled CoTs lead to the same answer cluster, $H_{\mathrm{sem}}(x)=0$.
When sampled CoTs lead to many different answer clusters, $H_{\mathrm{sem}}(x)$ becomes larger.
Thus, semantic entropy measures the disagreement among generated pseudo-CoTs at the answer level.

\subsection{Entropy Concentration and Majority Pseudo-Answers}

Semi-CoT uses the majority answer cluster as the pseudo answer.
Let
\begin{equation}
p_{\max}(x)=\max_{i}p_i
\end{equation}
be the empirical mass of the majority answer cluster.
The following lemma shows that low semantic entropy forces the answer distribution to concentrate on a dominant cluster.

\noindent\textbf{Lemma 1 (Entropy concentration).}
For any unlabeled question $x$, let $H_{\mathrm{sem}}(x)$ be defined by Eq.~\eqref{eq:theory_semantic_entropy}.
If $H_{\mathrm{sem}}(x)\leq \delta$, then
\begin{equation}
p_{\max}(x)
\geq
K^{-\delta}.
\label{eq:pmax_bound}
\end{equation}

\noindent\textit{Proof.}
Let $H(x)=-\sum_i p_i\log p_i$ be the unnormalized entropy.
Since $p_i\leq p_{\max}(x)$ for every $i$, we have
\begin{equation}
\log \frac{1}{p_i}
\geq
\log \frac{1}{p_{\max}(x)}.
\end{equation}
Therefore,
\begin{equation}
H(x)
=
\sum_i p_i\log \frac{1}{p_i}
\geq
\sum_i p_i\log \frac{1}{p_{\max}(x)}
=
\log \frac{1}{p_{\max}(x)}.
\end{equation}
This gives $p_{\max}(x)\geq \exp(-H(x))$.
Since $H_{\mathrm{sem}}(x)=H(x)/\log K$ and $H_{\mathrm{sem}}(x)\leq \delta$, we obtain
\begin{equation}
p_{\max}(x)\geq \exp(-\delta\log K)=K^{-\delta}.
\end{equation}
This completes the proof.
\hfill$\square$

Lemma 1 explains why the entropy gate behaves as an agreement filter.
If the entropy threshold $\delta$ is small, accepted pseudo-CoTs must have a concentrated answer distribution.
For example, when $K=3$ and $\delta<\log_K 2$, Eq.~\eqref{eq:pmax_bound} implies $p_{\max}(x)>1/2$.
In this case, the accepted pseudo answer is not only the largest cluster, but also a strict majority answer.

\subsection{Semantic Entropy and Pseudo-CoT Reliability}

Low entropy does not mathematically guarantee that the majority answer is correct.
A model can consistently produce the same wrong answer.
Therefore, we introduce a mild assumption connecting answer concentration and correctness.

\noindent\textbf{Assumption 1 (Semantic concentration--correctness).}
For an unlabeled question $x$, let $C^\star(x)$ denote the answer cluster that contains the correct answer.
Let $C_{\max}(x)$ denote the majority answer cluster.
The probability that the majority cluster is correct increases as the semantic entropy decreases:
\begin{equation}
\Pr[C_{\max}(x)=C^\star(x)\mid H_{\mathrm{sem}}(x)=h]
\quad
\text{is non-increasing in } h .
\label{eq:monotone_assumption}
\end{equation}

This assumption is consistent with the self-training view of CoT reasoning.
When independently sampled reasoning paths agree on the same answer, the model has lower semantic uncertainty.
When sampled paths disagree, the model has higher uncertainty.
Thus, low semantic entropy should increase the chance that the selected pseudo answer is reliable.

Under this assumption, the entropy gate improves the expected quality of pseudo-CoTs.

\noindent\textbf{Proposition 1 (Reliability of entropy-filtered pseudo-CoTs).}
Let $m(x)=\mathbb{I}[H_{\mathrm{sem}}(x)\leq\delta]$ be the Semi-CoT entropy gate.
Under Assumption 1, the expected pseudo-answer precision of accepted samples satisfies
\begin{equation}
\mathbb{E}
\left[
\mathbb{I}[C_{\max}(x)=C^\star(x)]
\mid
m(x)=1
\right]
\geq
\mathbb{E}
\left[
\mathbb{I}[C_{\max}(x)=C^\star(x)]
\right],
\label{eq:precision_improvement}
\end{equation}
when the accepted set has lower average semantic entropy than the full unlabeled set.

\noindent\textit{Proof sketch.}
Assumption 1 states that the correctness probability of the majority answer cluster is higher when semantic entropy is lower.
The entropy gate keeps only samples with $H_{\mathrm{sem}}(x)\leq\delta$.
If the accepted set has lower average entropy than the original unlabeled set, then its expected majority-cluster correctness is higher.
Thus, the accepted pseudo-CoTs have higher expected pseudo-answer precision than unfiltered pseudo-CoTs.
\hfill$\square$

This proposition explains the role of semantic entropy in Semi-CoT.
The gate does not prove correctness for every sample.
It improves the average reliability of the pseudo-CoT bank by removing high-disagreement questions.

\subsection{From Pseudo-Answer Reliability to Pseudo-CoT Reliability}

A pseudo-CoT contains both a reasoning chain and a final answer.
Even if the final answer is correct, the reasoning chain can still be invalid.
Therefore, answer-level entropy is an incomplete but useful proxy for pseudo-CoT reliability.
We formalize this using a decomposition of pseudo-CoT correctness.

Let $Z_y(x)$ denote the event that the selected pseudo answer is correct:
\begin{equation}
Z_y(x)=\mathbb{I}[\hat y=C^\star(x)].
\end{equation}
Let $Z_r(x)$ denote the event that the selected reasoning chain is valid:
\begin{equation}
Z_r(x)=\mathbb{I}[\hat r \text{ supports } \hat y].
\end{equation}
The pseudo-CoT is reliable only when both events hold:
\begin{equation}
Z_{\mathrm{cot}}(x)=Z_y(x)\cdot Z_r(x).
\end{equation}
Then the pseudo-CoT precision can be written as
\begin{equation}
\Pr[Z_{\mathrm{cot}}(x)=1]
=
\Pr[Z_y(x)=1]\,
\Pr[Z_r(x)=1\mid Z_y(x)=1].
\label{eq:cot_precision_decomp}
\end{equation}

Eq.~\eqref{eq:cot_precision_decomp} shows that answer consistency is necessary but not sufficient.
The entropy gate mainly improves $\Pr[Z_y(x)=1]$.
It does not fully verify $\Pr[Z_r(x)=1\mid Z_y(x)=1]$.
This explains why a future version of Semi-CoT should include reasoning-level verification, such as verifier scoring, step-level consistency, or symbolic checking.
The current report focuses on answer-level semantic entropy because it is simple, model-agnostic, and cheap to compute.

\subsection{Why Reliable Pseudo-CoTs May Still Cause Negative Transfer}

The pilot experiments show that high pseudo precision does not always lead to higher task accuracy.
This is because pseudo-CoT reliability and demonstration relevance are different.
A pseudo-CoT can be correct for its own question but still irrelevant or harmful for another test question.

Let $x$ be a test question.
Let $\mathcal{S}(x)=\{d_1,\ldots,d_M\}$ be the selected demonstration set from the pseudo reasoning bank.
Each demonstration $d_m=(x_m,\hat r_m,\hat y_m)$ has two properties:
its reliability and its relevance to $x$.
We define the event
\begin{equation}
R_m=\mathbb{I}[d_m \text{ is a reliable pseudo-CoT}],
\end{equation}
and the event
\begin{equation}
G_m(x)=\mathbb{I}[d_m \text{ is relevant to } x].
\end{equation}
The test prediction is most likely to improve when both events hold.

\noindent\textbf{Proposition 2 (Reliability--relevance decomposition).}
Let $\mathcal{E}(x)$ be the event that Semi-CoT predicts the wrong answer for test question $x$.
Then the error probability can be decomposed as
\begin{equation}
\Pr[\mathcal{E}(x)]
\leq
\Pr[\mathcal{E}(x)\mid \mathcal{R}(x),\mathcal{G}(x)]
+
\Pr[\neg \mathcal{R}(x)]
+
\Pr[\neg \mathcal{G}(x)],
\label{eq:relevance_bound}
\end{equation}
where $\mathcal{R}(x)=\bigcap_{m=1}^{M}R_m$ denotes that all selected demonstrations are reliable, and $\mathcal{G}(x)=\bigcap_{m=1}^{M}G_m(x)$ denotes that all selected demonstrations are relevant.

\noindent\textit{Proof.}
By the law of total probability,
\begin{equation}
\Pr[\mathcal{E}(x)]
=
\Pr[\mathcal{E}(x),\mathcal{R}(x),\mathcal{G}(x)]
+
\Pr[\mathcal{E}(x),\neg(\mathcal{R}(x)\cap\mathcal{G}(x))].
\end{equation}
The first term is upper bounded by
\begin{equation}
\Pr[\mathcal{E}(x)\mid \mathcal{R}(x),\mathcal{G}(x)].
\end{equation}
The second term is upper bounded by
\begin{equation}
\Pr[\neg(\mathcal{R}(x)\cap\mathcal{G}(x))]
\leq
\Pr[\neg\mathcal{R}(x)]+\Pr[\neg\mathcal{G}(x)].
\end{equation}
Combining the two bounds gives Eq.~\eqref{eq:relevance_bound}.
\hfill$\square$

Proposition 2 explains why Semi-CoT needs both an entropy gate and a demonstration retrieval strategy.
The entropy gate reduces $\Pr[\neg \mathcal{R}(x)]$ by filtering unreliable pseudo-CoTs.
A retrieval mechanism should reduce $\Pr[\neg \mathcal{G}(x)]$ by selecting demonstrations that match the test question.
If demonstrations are selected randomly, $\Pr[\neg \mathcal{G}(x)]$ can remain large.
This can cause negative transfer even when the pseudo-CoT bank has high precision.

\subsection{Implications for Semi-CoT}

The above analysis gives three conclusions.

First, semantic entropy controls answer concentration.
Low entropy implies that sampled CoT paths agree on a dominant answer cluster.
This justifies using semantic entropy as a first-stage pseudo-CoT selection signal.

Second, entropy filtering improves expected pseudo-answer precision under a natural concentration--correctness assumption.
This matches the empirical observation that the accepted pseudo-CoTs in our pilot experiments have high pseudo-answer precision.

Third, pseudo-CoT reliability alone is not enough.
A correct pseudo-CoT may still be an irrelevant demonstration for a test question.
Thus, Semi-CoT has two separate design problems:
\begin{equation}
\text{pseudo-CoT reliability}
\quad\text{and}\quad
\text{pseudo-CoT relevance}.
\end{equation}
The current entropy gate mainly addresses reliability.
Future work should improve relevance through stronger retrieval, reasoning-type matching, or student training.

Overall, the analysis supports the central motivation of Semi-CoT:
unlabeled questions can provide useful pseudo reasoning signals, but these signals must be selected and used carefully.
\section{Experiments}
\label{sec:experiments}

\subsection{Experimental Setup}

We conduct pilot experiments to evaluate whether unlabeled questions can provide reliable pseudo reasoning supervision.
The experiments are performed on four arithmetic reasoning datasets: AQuA, SVAMP, GSM8K, and MultiArith.
For each dataset, we construct a small labeled set and treat the remaining examples as unlabeled questions for pseudo-CoT generation.
Gold answers of unlabeled questions are not used for pseudo-CoT filtering.
They are used only for offline diagnosis of pseudo-answer precision.

We compare three inference methods.
\textbf{Zero-shot-CoT} uses the standard CoT trigger without any demonstrations.
\textbf{Zero-shot-CoT + SC} samples multiple CoT outputs and predicts the majority answer.
\textbf{Semi-CoT} first builds a pseudo reasoning bank from unlabeled questions and then uses accepted pseudo-CoTs as demonstrations.
Unless otherwise stated, Semi-CoT uses answer-level semantic entropy for pseudo-CoT filtering.
A pseudo-CoT is accepted when its semantic entropy is below the threshold $\delta$.

The current experiments are pilot studies before full-set evaluation.
AQuA is evaluated on 50 examples, while SVAMP, GSM8K, and MultiArith are evaluated on 100 examples.
We use a 10\% labeled ratio for pseudo-CoT construction.
The main goal of this section is to test whether entropy-filtered pseudo-CoTs are reliable and whether they can provide initial performance gains.

\subsection{Pilot Accuracy Results}

Table~\ref{tab:pilot_accuracy} reports the pilot accuracy of Zero-shot-CoT, Zero-shot-CoT with self-consistency, and Semi-CoT.
Semi-CoT improves over Zero-shot-CoT by $2$ points on SVAMP and $1$ point on GSM8K.
On AQuA, Semi-CoT is lower than Zero-shot-CoT, suggesting mild negative transfer.
On MultiArith, all methods reach $100\%$, indicating a ceiling effect in this pilot setting.

\begin{table}[t]
\centering
\small
\caption{
\textbf{Pilot accuracy results.}
We report accuracy on AQuA-50, SVAMP-100, GSM8K-100, and MultiArith-100.
Semi-CoT improves on SVAMP and GSM8K, while AQuA shows negative transfer and MultiArith reaches a ceiling.
}
\label{tab:pilot_accuracy}
\begin{tabular}{lcccc}
\toprule
Dataset & Size & Zero-shot-CoT & + SC & Semi-CoT \\
\midrule
AQuA & 50 & 88.0 & 86.0 & 86.0 \\
SVAMP & 100 & 90.0 & 91.0 & 92.0 \\
GSM8K & 100 & 89.0 & 89.0 & 90.0 \\
MultiArith & 100 & 100.0 & 100.0 & 100.0 \\
\bottomrule
\end{tabular}
\end{table}

The results show that Semi-CoT has a positive but still limited signal.
The improvement is not large enough to claim significant performance gains.
However, the results are sufficient to justify full-set experiments on SVAMP and GSM8K, where the pilot results show non-zero improvements.
MultiArith is less informative at this stage because all methods already reach perfect accuracy.

\subsection{Pseudo-CoT Reliability Diagnosis}

Table~\ref{tab:pseudo_diagnosis} reports pseudo-CoT filtering statistics.
The entropy gate accepts 35 out of 45 unlabeled AQuA questions, 81 out of 90 unlabeled SVAMP questions, 81 out of 90 unlabeled GSM8K questions, and all 90 unlabeled MultiArith questions.
The accepted pseudo-CoTs have high pseudo-answer precision across all datasets.
The precision is $97.14\%$ on AQuA, $96.30\%$ on SVAMP, $91.36\%$ on GSM8K, and $100.00\%$ on MultiArith.

\begin{table}[t]
\centering
\small
\caption{
\textbf{Pseudo-CoT reliability diagnosis.}
Gold answers of unlabeled questions are not used during pseudo-CoT filtering.
They are used only for offline diagnosis.
The entropy gate selects high-precision pseudo-CoTs across datasets.
}
\label{tab:pseudo_diagnosis}
\begin{tabular}{lccccc}
\toprule
Dataset & Labeled & Unlabeled & Accepted & Accept Rate & Pseudo Precision \\
\midrule
AQuA & 5 & 45 & 35/45 & 77.78 & 97.14 \\
SVAMP & 10 & 90 & 81/90 & 90.00 & 96.30 \\
GSM8K & 10 & 90 & 81/90 & 90.00 & 91.36 \\
MultiArith & 10 & 90 & 90/90 & 100.00 & 100.00 \\
\bottomrule
\end{tabular}
\end{table}

These results support the use of semantic entropy as a pseudo-CoT reliability signal.
However, high pseudo-answer precision does not automatically translate into higher task accuracy.
For example, AQuA obtains high pseudo precision but still shows negative transfer.
This indicates that pseudo-CoT reliability and demonstration relevance are different factors.
The entropy gate controls the reliability of the pseudo reasoning bank, while the demonstration selection strategy controls whether the selected pseudo-CoTs are useful for a given test question.

\subsection{Effect of Demonstration Retrieval}

To test whether demonstration relevance improves Semi-CoT, we compare random demonstration selection with TF-IDF retrieval.
The TF-IDF retriever is implemented in pure Python without additional dependencies.
For each test question, it retrieves pseudo-CoTs whose questions have the highest lexical similarity to the test question.
The current test question is automatically removed from the pseudo bank to avoid data leakage.

Table~\ref{tab:retrieval} shows the preliminary comparison.
TF-IDF retrieval performs the same as random selection in the current pilot subsets.
This suggests that simple lexical similarity is not sufficient to improve pseudo-CoT usage.
A stronger retrieval strategy may require semantic embeddings, answer-type matching, or reasoning-pattern matching.

\begin{table}[t]
\centering
\small
\caption{
\textbf{Random versus TF-IDF demonstration selection.}
TF-IDF retrieval does not improve over random selection in the current pilot subsets.
This suggests that lexical similarity alone is not enough for reliable pseudo-CoT usage.
}
\label{tab:retrieval}
\begin{tabular}{lccc}
\toprule
Dataset & Size & Semi-CoT-random & Semi-CoT-TF-IDF \\
\midrule
AQuA & 50 & 84.0 & 84.0 \\
SVAMP & 100 & 92.0 & 92.0 \\
GSM8K & 100 & 90.0 & 90.0 \\
\bottomrule
\end{tabular}
\end{table}

\subsection{Summary of Findings}

The pilot experiments give three observations.
First, entropy filtering selects high-precision pseudo-CoTs across all evaluated datasets.
This supports the central motivation that unlabeled questions can provide reliable pseudo reasoning signals.
Second, accuracy improvements are currently small.
Semi-CoT improves on SVAMP and GSM8K, but AQuA shows negative transfer and MultiArith has a ceiling effect.
Third, random and TF-IDF demonstration selection are not sufficient to fully exploit the pseudo reasoning bank.
Therefore, the next stage should focus on full-set evaluation for SVAMP and GSM8K, multi-seed runs, and stronger retrieval or student training.

Based on the current pilot results, the most informative full-set experiments are SVAMP and GSM8K.
MultiArith is not prioritized because the pilot subset already reaches $100\%$ accuracy.
AQuA should be retained as a stress test for negative transfer.

\section{Conclusions}
\label{sec:conclusion}

In this technical report, we revisited chain-of-thought reasoning under limited supervision and introduced \textbf{Semi-supervised Chain-of-Thought Learning}.
The central idea is to move CoT beyond an inference-time prompting technique and treat model-generated reasoning chains as potential pseudo-supervision for unlabeled questions.
Motivated by the connection between CoT reasoning and self-training, we formulated unlabeled reasoning supervision as
\[
x_u \rightarrow (\hat r_u,\hat y_u),
\]
where an unlabeled question is converted into a pseudo reasoning chain and a pseudo answer.

We proposed \textbf{Semi-CoT}, a minimal framework for this setting.
Semi-CoT samples multiple reasoning paths for each unlabeled question, estimates answer-level semantic entropy, and accepts only low-entropy pseudo-CoTs into a pseudo reasoning bank.
The accepted pseudo-CoTs are then used as demonstrations during inference.
This simple implementation does not fine-tune model parameters, but it directly tests whether unlabeled questions can provide reliable pseudo reasoning signals.

Pilot experiments on AQuA, SVAMP, GSM8K, and MultiArith show that semantic-entropy filtering selects high-precision pseudo-CoTs.
The accepted pseudo-CoTs achieve pseudo-answer precision from $91.36\%$ to $100.00\%$ across datasets.
In terms of accuracy, Semi-CoT gives small gains on SVAMP and GSM8K, while AQuA shows negative transfer and MultiArith reaches a ceiling effect.
These results suggest that unlabeled questions can indeed provide reliable pseudo reasoning signals, but reliable pseudo-CoT selection alone is not sufficient for consistent performance improvement.

The current study also reveals an important limitation.
Pseudo-CoT reliability and demonstration relevance are different problems.
A pseudo-CoT may be correct for its own question but irrelevant or even harmful for another test question.
Our TF-IDF retrieval pilot shows that simple lexical retrieval does not yet improve over random demonstration selection.
Therefore, future work should study stronger retrieval mechanisms, reasoning-pattern matching, verifier-based filtering, and training-based student adaptation.

Overall, this report establishes Semi-supervised Chain-of-Thought Learning as a new problem setting.
The main conclusion is conservative but useful:
unlabeled questions can provide reliable pseudo reasoning supervision, but effective Semi-CoT systems must jointly address pseudo-CoT reliability, demonstration relevance, and student learning.
% \section*{Acknowledgements}

% \clearpage

\bibliographystyle{plainnat}
\setlength{\bibhang}{0pt}
\setlength\bibindent{0pt}
\bibliography{main}

\clearpage

\clearpage
\appendix
\startcontents[chapters]
\setcounter{page}{1}

% \begin{center}
%   \textbf{\Large StableVLA: Towards Robust Vision-Language-Action Models without Extra Data} \vspace{0.5cm} \\
%   {\Large Appendix}
%   \vspace{0.5cm}
% \end{center}

% \printcontents[chapters]{}{1}{}

% \input{sections/Suppl/appendix_a}
% \input{sections/Suppl/appendix_b}
% \input{sections/Suppl/appendix_c}
% \input{sections/Suppl/appendix_d}
% \input{sections/Suppl/appendix_e}

\end{document}